\documentclass[11pt]{article}

\usepackage[a4paper,margin=1in]{geometry}
\usepackage{amsmath}
\usepackage{amssymb}
\usepackage{graphicx}
\usepackage[super,comma,sort&compress]{natbib}
\usepackage[colorlinks=true,allcolors=blue]{hyperref}
\usepackage{orcidlink}
\usepackage{authblk}
\usepackage{jabbrv} 

\makeatletter
\def\bibinfo#1#2{\def\jabbrv@field{#1}\def\jabbrv@journal{journal}%
  \ifx\jabbrv@field\jabbrv@journal\JournalTitle{#2}\else#2\fi}
\makeatother

\DefineJournalAbbreviation{Bioinformatics}{Bioinform}
\DefineJournalAbbreviation{Medicine}{Med}
\DefineJournalAbbreviation{Royal}{R}
\DefineJournalAbbreviation{Applied}{Appl}

\DeclareMathOperator*{\argmax}{argmax}

\title{\vspace{-1.5em}\textbf{Reliable data clustering\\ with Bayesian community detection}\vspace{-0.5em}}

\author[1]{Magnus Neuman\,\orcidlink{0000-0002-3599-9374}\thanks{Corresponding author: magnus.neuman@umu.se}}
\author[1]{Jelena Smiljani\'c\,\orcidlink{0000-0003-0124-1909}}
\author[1]{Martin Rosvall\,\orcidlink{0000-0002-7181-9940}}

\affil[1]{Integrated Science Lab, Department of Physics, Umeå University, SE-90187 Umeå, Sweden}

\date{}

\begin{document}

\maketitle

\begin{abstract}
\noindent
From neuroscience and genomics to systems biology and ecology, researchers rely on clustering similarity data to uncover modular structure. Most methods work in two steps: first reduce noise, then cluster. The first step runs without regard to the second: it forces arbitrary choices, discards useful information, and can impose structure where none exists. Here we bring the two steps together. Using established Bayesian community detection methods, we let one minimum description length criterion choose both how much to sparsify the data and how to cluster it. This one-step approach keeps all the samples for inference, removes the arbitrary threshold choice, and returns a single cluster when there is no real structure. On synthetic data, our approach recovers clusters under more noise and with fewer samples than standard methods. On gene co-expression data from the human brain, it finds more coherent and more stable clusters than WGCNA, the standard tool in the field. Uniting sparsification and clustering in a single model-selection step provides a principled, noise-resistant way to uncover modular structure in similarity data across fields. 
\end{abstract}

\section*{Introduction}

Researchers apply clustering techniques to analyze similarity data from diverse systems \cite{Everitt2011cluster}, such as the human brain \cite{Bullmore, Hawrylycz}, ecological assemblages \cite{Barberan, calatayud_elife, deVries}, the regulatory mechanisms of genes \cite{eisen1998cluster,Wang}, and plant-pathogen interactions \cite{Mishra2021}.
Despite their diversity, the common goal is to partition observed features, such as species or genes, into clusters based on their similarity estimated from repeated samples, revealing meaningful structure and underlying organization. For instance, species co-occurrence data can reveal species groups sharing similar environmental conditions, and gene expression patterns can reveal clusters of genes with similar functions.

The sheer number of pairwise associations in high-dimensional data threatens to drown signal in noise \cite{wainwright2019high}. When researchers apply clustering methods directly to unfiltered data or similarity matrices, noise, outliers, and arbitrary parameters obscure the clusters \cite{Huber1981robust, Altman_clustering_2017}. Current approaches tackle this challenge in two disconnected steps: first reduce noise, then find clusters \cite{vzurauskiene2016pcareduce,allaoui2020considerably,hozumi2021umap}. This separation creates fundamental problems.

One class of two-step methods first compresses high-dimensional data into fewer dimensions, often with PCA \cite{jolliffe2002pca}
or UMAP \cite{mcinnes2018umap},
then applies clustering algorithms such as $k$-means \cite{hartigan1979algorithm,hozumi2021umap} or hierarchical clustering \cite{vzurauskiene2016pcareduce}.
This two-step process creates multiple problems. First, researchers must make arbitrary choices twice: how many dimensions to keep and how many clusters to find. Second, the dimensionality reduction may discard information that defines clusters while preserving variance that reflects noise \cite{Huber1981robust, yeung2001principal,chen2020robust}. Third, these methods can detect clusters even when there is no modular structure in the underlying data \cite{Altman_clustering_2017}. The root problem is that the reduction step determines what the clustering step can possibly find, yet operates blind to the clustering objective.

Another class of two-step methods first sparsifies correlation matrices into networks through thresholding, then applies clustering algorithms \cite{Horvath, VANDENHEUVEL2008, Bleker2024}. Researchers compute pairwise correlations between all features, typically using Pearson correlation. Accurately estimating all these correlations requires many samples relative to the number of features \cite{Ledoit2004}. This requirement rarely holds in practice. For example, gene co-expression studies demand expensive and time-consuming gene sequencing experiments \cite{sboner2011real}, while ecologists contend with detection errors and resource constraints when collecting plant occurrence data across field plots \cite{milberg2008observer}. With scarce data yielding unreliable correlations, researchers sparsify these noisy matrices into networks: features form nodes and correlations above a chosen threshold form links \cite{Masuda}.
While this thresholding to reduce noise enables network algorithms, including community detection methods for clustering, the sparsification optimizes for networks, not clusters.

Finding accurate links and finding reliable clusters require different objectives \cite{Perkins2209, MacMahon2015, Zahor2016, Neuman}, yet most clustering pipelines apply thresholding without theoretical justification, relying instead on various heuristic network criteria \cite{Bleker2024, Jay2012}.
Each threshold creates a different network: lower thresholds retain noise in dense networks; higher thresholds remove signal in sparse networks.
Even the all-in-one package WGCNA (Weighted Gene Correlation Network Analysis), with its soft thresholding based on scale-free topology \cite{Horvath,Langfelder}, optimizes network accuracy before clustering.
When sparsification predetermines what clustering can detect, unreliable correlations can produce distorted clusters.

Network and cluster optimization diverge because accurate links and reliable clusters require different trade-offs. Triangles illustrate this link-cluster trade-off: when two features appear related simply because they both correlate with a third, graphical lasso \cite{Friedman2007, Yuan} and other partial correlation methods \cite{Meinshausen, Zou_elasticnet} identify these indirect associations and remove them to improve network accuracy. For our clustering objective, such network-focused precision adds no value. Triangle closure can preserve or even enhance modular structure by adding more links within than between clusters. We previously demonstrated that cross-validation based on cluster quality instead of network accuracy produces networks that better reveal modular structure, whether applied directly to correlation networks \cite{Neuman} or to graphical lasso \cite{Neuman2}. These cluster-focused optimizations succeed because mesoscale clusters remain robust across reasonable threshold ranges when genuine modular structure exists: clusters tolerate both missing and spurious links within bounds \cite{Augustson1970, Rosvall_mappingchange, smiljanic1, smiljanic2, Neuman}, so link-level accuracy is not the right criterion for cluster detection. Cross-validation targets the right objective, but splits already scarce data.

Current methods make arbitrary parameter choices or waste samples on validation. We lack a one-step method that applies principled model selection to similarity data without discarding samples. Here we present a single minimum description length (MDL) criterion that uses established Bayesian community detection methods to select the sparsification threshold, the number of clusters, and the cluster assignments together, using every sample for inference. Because this criterion must work at the edge of what the data can support, we first ask when correlation data contain enough signal for clusters to be detectable at all.

To establish this limit of reliable cluster detection, we generate synthetic data from a planted block structure and analyze Pearson correlation.
Its analytically tractable sampling distribution lets us characterize when within- and between-cluster correlation distributions separate enough for clusters to become detectable as a function of sample size, population correlation strength, number of clusters, and number of features. Pearson correlation also remains the most widely used similarity measure in practice \cite{zitnik2024}.
Other similarity measures, including field-specific measures such as molecular or species similarity \cite{Amoroso, Leinster}, require their own detectability analysis, but we expect qualitatively similar behavior.

Given a network, Bayesian community detection methods \cite{smiljanic1,smiljanic2, peixotoPRE2017,Pexioto22TC} use the MDL principle to balance model complexity against goodness of fit \cite{Rissanen}, determining both the number of clusters and the node assignments.
We let the same criterion select the sparsification threshold, avoiding both collapsed clusters in dense networks and fragmented clusters in sparse networks. This one-step approach offers three advantages:
it replaces two arbitrary choices---threshold value and number of clusters---with MDL-based compression as one principled criterion;
it preserves all samples for inference, rather than discarding data through cross-validation;
and it avoids imposing structure on noise, returning a single cluster when no modular structure exists.

We evaluate two Bayesian methods for community detection: the Degree-Corrected Stochastic Block Model (DC-SBM) \cite{peixotoPRE2017} and the Regularized Map Equation \cite{smiljanic2} implemented in Infomap \cite{MapEqRev}, both widely used across scientific disciplines. They operate directly on the network and require no model for the underlying similarity measure, making them applicable beyond Pearson correlation. The DC-SBM offers a probabilistic framework, while the Regularized Map Equation is based on an information-theoretic criterion with a Bayesian prior to mitigate overfitting.

On synthetic data, the Regularized Map Equation recovers planted partitions in higher-noise settings than the DC-SBM, cross-validation, and hierarchical clustering. Compared with WGCNA on gene co-expression data, it identifies more functionally coherent and stable clusters.

\section*{Results}

\subsection*{Correlation distributions}

\begin{figure*}[tb]
\centering
\includegraphics[width=\textwidth]{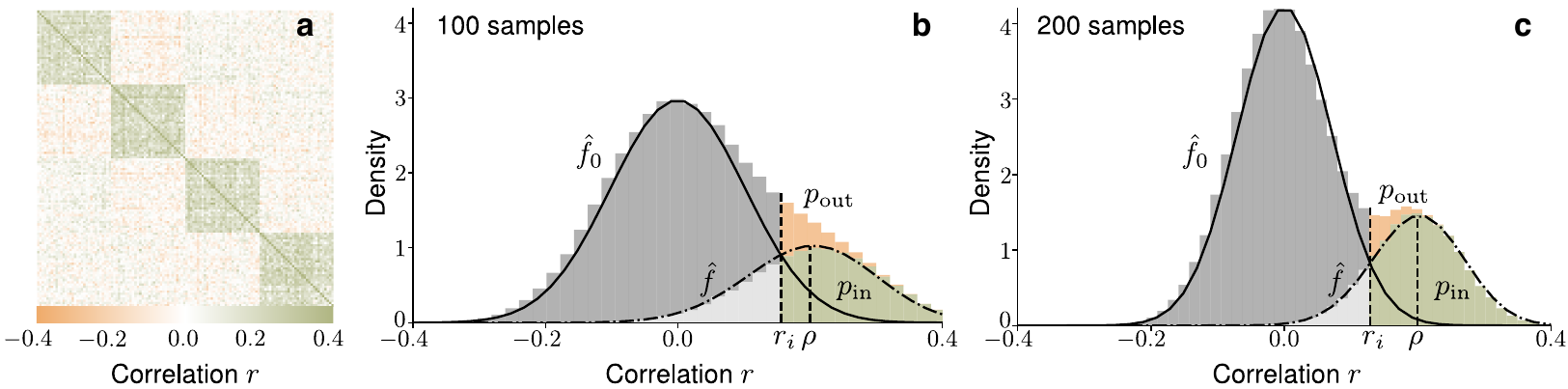}
\caption{\label{fig:1}\textbf{Overlapping correlation distributions.} In \textbf{a}, the planted structure is visible in the observed correlation matrix $\hat{\Sigma}$, since the within-cluster correlations on average exceed the between-cluster correlations. This separation depends on the number of samples, nodes, and clusters, and on the population correlation $\rho$. Colorbar truncated at 0.4; diagonal entries have correlation 1. With few samples in \textbf{b}, the within- and between-cluster distributions $\hat{f}$ and $\hat{f}_0$ overlap, and the number of false positives using the threshold $r_i$, where $\hat{f}_0$ and $\hat{f}$ intersect, is relatively large, making correct inference difficult. With more samples in \textbf{c}, the situation becomes more favorable. The fractions of within- and between-cluster correlations above $r_i$ are $p_{\mathrm{in}}$ (green, true positives) and $p_{\mathrm{out}}$ (orange, false positives) respectively.}
\end{figure*}

To assess the detectability of modular structure in correlational data, we first analyze synthetic datasets with known cluster structure.
We generate synthetic data using a stochastic block model adapted for correlational data, with $N$ observed features in $q$ planted clusters of equal size. To control the noise level, we vary both the strength of the planted population correlation $\rho$ between features in the same cluster and the number of samples $L$.
We sample data $X\in \mathbb{R}^{L\times N}$ from a multivariate normal distribution, such that each row $X_i\sim \mathcal{N}(0,\Sigma)$, where the covariance matrix $\Sigma$ has block structure:
\begin{equation}
\Sigma_{ij} = \begin{cases}
1, & \text{if } i = j, \\
\rho, & \text{if } i\neq j, \text{ and } i \text{ and } j \text{ belong to the same cluster}, \\
0, & \text{otherwise}.
\end{cases}
\end{equation}
The matrix $\hat{\Sigma}$ 
is the Pearson correlation matrix of $X$ and serves as input for our cluster inference. With enough samples and strong population correlation, this matrix reveals the planted structure (Fig.~\ref{fig:1}a). The number and sizes of clusters also influence detectability: small clusters cover a smaller portion of the covariance matrix and yield fewer within-cluster correlation samples, producing noisier estimates of the within-cluster correlation distribution.

To clarify the inference problem, we consider the distribution of correlation $r$, which has a known analytical expression \cite{Hotelling}.
The observed correlation distribution is a superposition of the within-cluster distribution $\hat{f}(r; \rho, L, q, N)$ and the between-cluster distribution $\hat{f}_0(r; L, q, N)$ (see Methods for a derivation). These distributions depend weakly on the number of features $N$. Only when $N$ is small or approaches the number of clusters does this dependence become relevant. Figures~\ref{fig:1}b and c show these distributions for two different sample sizes. When the sample size is small, the overlap between $\hat{f}$ and $\hat{f}_0$ is large. More samples separate the distributions and enable correct inference of the underlying clusters.

Thresholding the observed correlations to separate signal from noise trades off true and false positives. The intersection $r_i$ between $\hat{f}(r; \rho, L, q, N)$ and $\hat{f}_0(r; L, q, N)$ provides a candidate optimal threshold (analytical derivation in Methods). A lower threshold adds more between-cluster than within-cluster correlations (more false than true positives); a higher threshold removes more within-cluster than between-cluster correlations (more true than false positives), as illustrated in Fig.~\ref{fig:1}b and c.

\begin{figure*}[tb]
\centering
\includegraphics[width=0.8\textwidth]{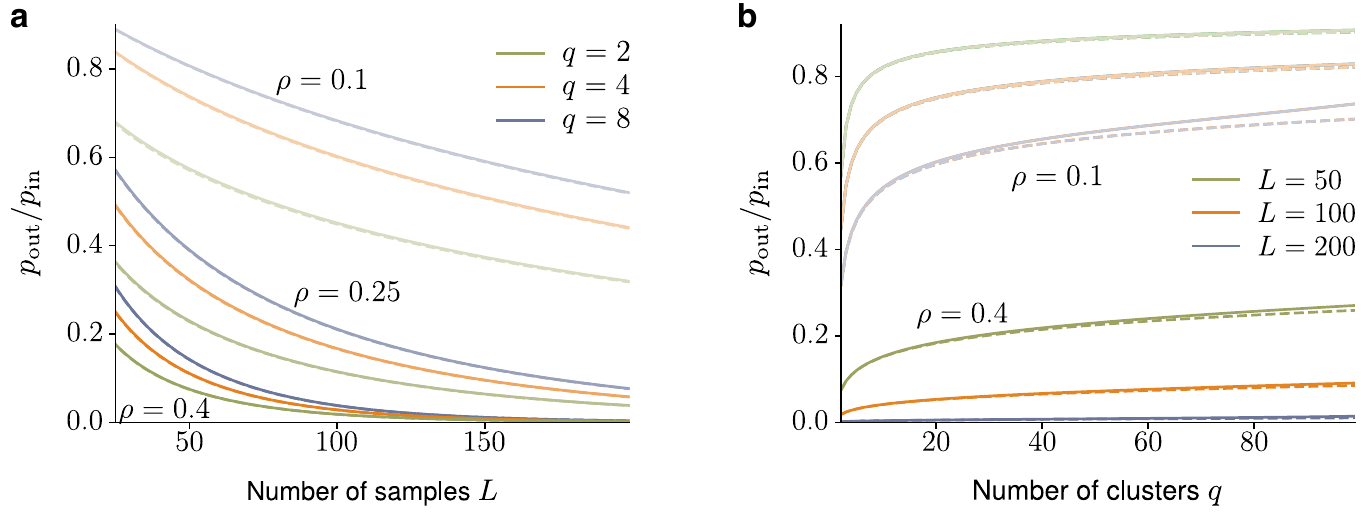}
\caption{\label{fig:2}\textbf{The transition to separated correlation distributions.} In \textbf{a}, the ratio $p_{\mathrm{out}}/p_{\mathrm{in}}$ shows the non-linear transition from overlapping to separated between- and within-cluster distributions as the number of samples $L$ increases. More samples are required if the number of clusters $q$ increases, particularly if the population correlation is weak. The number of features has a negligible effect, as shown by the barely visible dashed lines where the number of features is doubled. In \textbf{b}, splitting the network into more clusters increases $p_{\mathrm{out}}/p_{\mathrm{in}}$, making correct inference more difficult, particularly with weak population correlation. The number of features matters weakly in this case, where more features actually ease inference.}
\end{figure*}

This threshold gives the probabilities for within-cluster links
\begin{equation}
    p_{\mathrm{in}}(\rho, L, q, N) = \int_{r_i}^1 \hat{f}(r;\rho,L, q, N) dr
\end{equation}
and between-cluster links
\begin{equation}
    p_{\mathrm{out}}(L, q, N) = \int_{r_i}^1 \hat{f}_0(r;L, q, N) dr.
\end{equation}
The ratio $p_{\mathrm{out}}/p_{\mathrm{in}}$ reveals the transition from overlapping to separated distributions.
Figure~\ref{fig:2} shows this transition as the number of samples increases (Fig.~\ref{fig:2}a) and as the number of clusters increases (Fig.~\ref{fig:2}b). The transition is non-linear in both samples $L$ and clusters $q$. With a strong population correlation, $p_{\mathrm{out}}/p_{\mathrm{in}}$ decreases sharply as samples increase, while the decrease is more gradual with a weaker population correlation. With weak population correlation, more granular clusters sharply increase $p_{\mathrm{out}}/p_{\mathrm{in}}$ and raise the noise level (Fig.~\ref{fig:2}b), making correct inference more difficult. More features help here, moderately reducing noise. With many samples and strong population correlation, the number of clusters matters less.

From a practical perspective, if a group of features in the studied system correlates, detection improves non-linearly with the number of samples. This improvement depends weakly on the number of observed features $N$ and can even strengthen with more features. Fine-grained cluster structure is harder to detect than coarse structure in a system of the same size.

\subsection*{Bayesian cluster-based regularization}
With sample size established as a key limiting factor, we capitalize on Bayesian community detection methods with built-in regularization to make better use of limited data.
We consider two methods: the Regularized Map Equation \cite{smiljanic1, smiljanic2} and the Degree-Corrected Stochastic Block Model (DC-SBM) \cite{peixotoPRE2017}. These methods incorporate prior information about network parameters and structure to obtain a posterior estimate of network communities, but use different approaches.

The Regularized Map Equation builds on the map equation framework, which models flow on a network as a random walk \cite{RosvallPNAS2008, Edler1, MapEqRev}. Following the MDL principle, the optimal partition gives the shortest description of the random walker's path. This framework performs well in community detection benchmarks \cite{Lancichinetti, Aldecoa2013, tandon2021}. The Regularized Map Equation estimates posterior transition rates of the random walker using a Dirichlet prior on transition rates (see Methods). The Dirichlet prior prevents overfitting modular structure in sparse networks and spuriously inferring communities in random networks, two limitations of the standard map equation \cite{Lancichinetti, FORTUNATO2016}.

The DC-SBM uses a combinatorial approach, fitting a block model to the observed network data to minimize the description length of the partitioned network, also following the MDL principle.
Both the Regularized Map Equation and the DC-SBM infer an optimal partition of the network into communities with the corresponding description length. Neither method requires the number of communities to be known beforehand.

\begin{figure*}[tb]
\centering
\includegraphics[width=\textwidth]{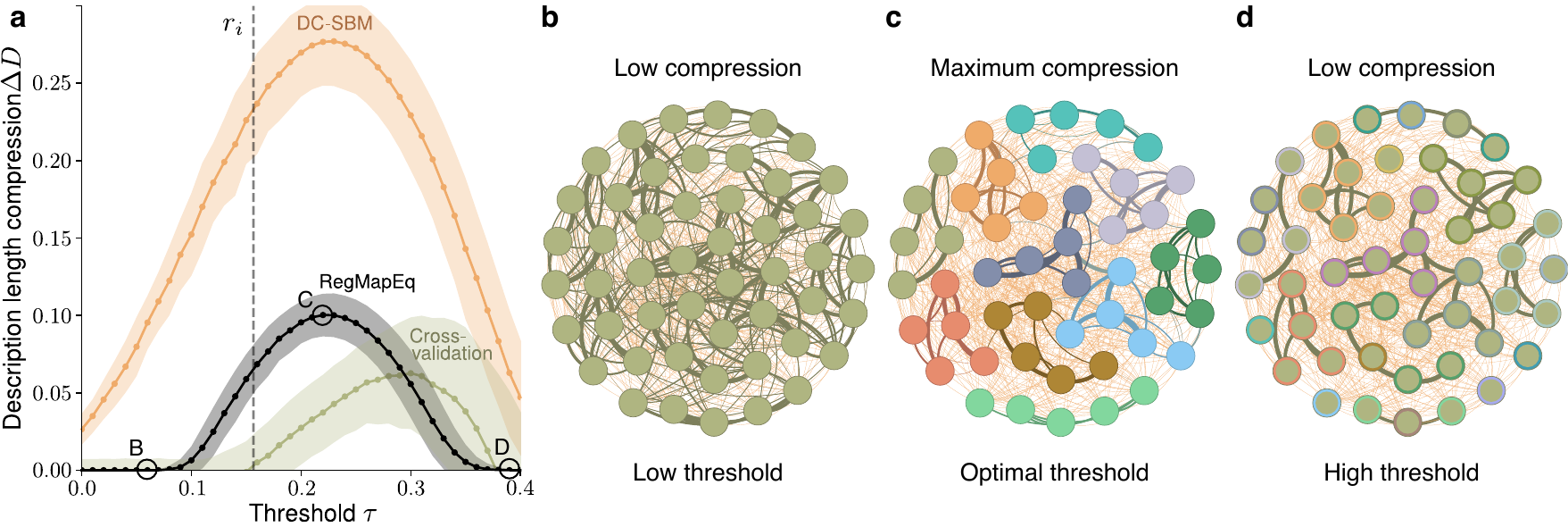}
\caption{\label{fig:SBM_IM}\textbf{Bayesian community detection and description length compression.} In \textbf{a}, the description length compression peaks to the right of the intersection $r_i$ of within- and between-cluster correlation distributions. The compression peak lies between the dense (\textbf{b}) and sparse (\textbf{d}) networks, where the signal of modular structure is maximized (\textbf{c}).
To avoid overfitting modular structure in sparse networks, Bayesian community detection uses a prior shown as orange, thin links in \textbf{b}--\textbf{d}. This approach regularizes correlation networks without splitting already scarce data. The node border colors in network \textbf{d} show the clusters without the prior.}
\end{figure*}

The MDL principle penalizes complexity and prevents spurious structure from emerging in noise.
Both methods are applied to the correlation network thresholded at $\tau$. The DC-SBM uses the resulting weighted network, retaining correlation values above the threshold, whereas the Regularized Map Equation operates on the corresponding unweighted network, as required by its prior (see Methods for details).
We define the description length compression $\Delta D$ at threshold $\tau$ as:
\begin{equation}
    \Delta D(\tau) = \frac{D^1(\tau) - D^*(\tau)}{D^1(\tau)},
\end{equation}
where $D^1(\tau)$ is the description length of an uncompressed model with all nodes in the same cluster, and $D^*(\tau)$ is the description length of the best partition inferred at threshold $\tau$, computed according to the MDL formulations of the Regularized Map Equation \cite{RosvallPNAS2008, MapEqRev} and the DC-SBM \cite{peixotoPRE2017}, respectively. When the threshold is low, the network is dense and shows no recoverable modular structure, resulting in low compression. At high thresholds, the network becomes sparser, increasing the risk of overfitting. Both methods guard against this overfitting: the Regularized Map Equation through its prior on network structure (Fig.~\ref{fig:SBM_IM}), and the DC-SBM through the increased information required to describe a more granular partition of the sparse network.

The MDL-optimal threshold $\tau^*$ lies between the sparse and dense extremes, where the description length compression peaks and the modular signal is strongest:
\begin{equation}
    \tau^* = \argmax_\tau \Delta D (\tau).
\end{equation}
Figure~\ref{fig:SBM_IM}a shows the compression $\Delta D$ as a function of $\tau$, comparing the Regularized Map Equation, the DC-SBM, and cross-validation as described in refs.~\citenum{Neuman, Neuman2}. We averaged results over 100 runs with $L=100$ samples, population correlation $\rho = 0.2$, $q=4$ clusters, and $N=256$ nodes. The compression is zero at both extremes of the threshold range. The peak compression occurs at $\tau^*\approx 0.22$ for both the Regularized Map Equation and the DC-SBM, identifying the optimal threshold for cluster detection.
The standard deviation is larger with cross-validation due to reduced sample size from data splitting, while the Regularized Map Equation and the DC-SBM use all available samples.

The MDL-optimal threshold $\tau^*$ exceeds the intersection threshold $r_i$. While $r_i$ balances false and true positives at the level of individual correlations, $\tau^*$ minimizes the description length of the inferred partition. False positives connect nodes in different clusters and reduce modular compressibility, increasing the description length for both methods. In contrast, removing some true positives within clusters often leaves the modular structure intact and incurs only a small penalty in description length. Link-level critiques of thresholding therefore do not apply when the objective is cluster detection: avoiding false positives justifies removing additional true positives, leading to $\tau^* > r_i$. Because cross-validation splits the data, it widens the between-cluster correlation distribution and requires a more stringent threshold to prevent false positives.

\begin{figure*}[tb]
\centering
\includegraphics[width=\textwidth]{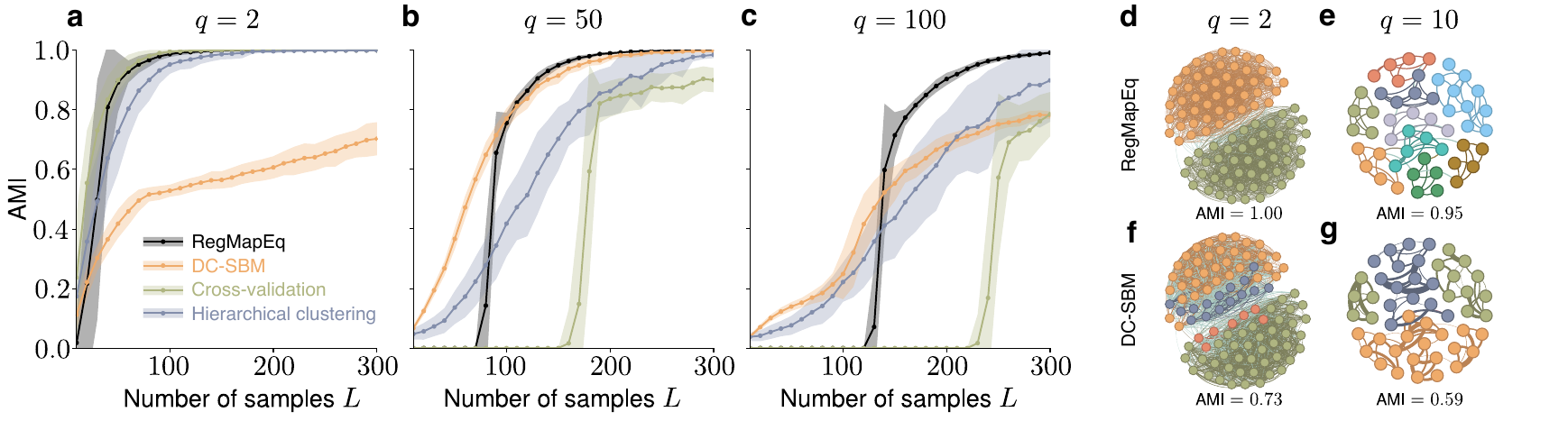}
\caption{\label{fig:AMI}\textbf{Method performance measured by the adjusted mutual information (AMI) between planted and inferred partitions.} For networks with $N=1000$ nodes, correlation strength $\rho=0.2$, and $q=2$ clusters, the noise level is low and all methods but the DC-SBM correctly infer the planted partition using few samples (\textbf{a}).
The DC-SBM assumes conditional link independence given the block structure and does not explicitly model triangle closure, whereas correlation networks tend to exhibit high transitivity.
With $q=50$ clusters, the Regularized Map Equation and the DC-SBM recover the planted partition with fewer samples than cross-validation and hierarchical clustering (\textbf{b}). With $q=100$ clusters, the noise level is higher, and hierarchical clustering struggles to infer the correct partition, as does the DC-SBM due to its resolution limit (\textbf{c}).
Hierarchical clustering and the DC-SBM infer clusters (AMI $>$ 0) even in high noise (\textbf{b} and \textbf{c} with few samples), making interpretation difficult. For schematic networks with $L=100$ samples, correlation strength $\rho=0.3$, and two 50-node clusters (\textbf{d}, \textbf{f}) or ten 5-node clusters (\textbf{e}, \textbf{g}), the DC-SBM overfits the large clusters due to high clustering coefficients (\textbf{f}) and underfits the small clusters due to its resolution limit (\textbf{g}).}
\end{figure*}

Figure~\ref{fig:AMI}a--c shows the adjusted mutual information (AMI), a commonly used information-theoretic measure of similarity between two partitions \cite{VinhAMI}, between planted and inferred partitions as a function of the number of samples $L$.
Both the Regularized Map Equation and the DC-SBM recover the planted partition with fewer samples than cross-validation, illustrating the advantage of Bayesian regularization in data-scarce settings.
As a reference, hierarchical clustering requires more samples to reliably infer the planted structure (details in Methods).
In the high-noise scenarios with $q = 50$ and $q = 100$ clusters (Fig.~\ref{fig:AMI}b and c), hierarchical clustering performs poorly since there is no clear signal of where to cut the dendrogram (Supplementary Fig.~1). Hierarchical clustering infers modular structure even when the data are close to pure noise. The DC-SBM exhibits similar behavior, giving AMI $> 0$ in noise-dominated settings.
The Regularized Map Equation detects no modular structure (AMI = 0) until the sample size provides sufficient signal, then recovers the planted clusters with high accuracy as samples increase.

The DC-SBM faces two challenges when working with correlational data.
First, its resolution limit causes small clusters and isolated nodes to merge into larger blocks (Fig.~\ref{fig:AMI}g) \cite{Peixoto2013_reslim}, leading to poor recovery in the $q=100$ setting where clusters are small. The nested DC-SBM does not resolve this challenge (Supplementary Fig.~2).
Second, the DC-SBM assumes conditional independence of links given the block structure, an assumption that correlation networks tend to violate through triangle closure \cite{Langford2001}, leading to overfitting in networks with high clustering coefficients \cite{Pexioto22TC}. In the example shown in Fig.~\ref{fig:AMI}f, where two 50-node clusters contain many triangles, the DC-SBM splits clusters into smaller ones.
Extending the DC-SBM to explicitly model triangle closure improves estimation of the clustering coefficient \cite{Pexioto22TC}, but does not resolve this overpartitioning in dense correlation networks (Supplementary Fig.~3).

We conclude that cluster inference depends primarily on sample size, population correlation, cluster count, and only weakly on the number of features. When samples are few, correlations weak, or clusters numerous, the distribution of between-cluster correlations overlaps with the within-cluster correlation distribution and correct inference becomes impossible. This detectability problem relates to the detectability limit of network communities, a well-studied limit in network science \cite{Decelle:PhysRevLett, abbe2016, fortunato2007}. For large and locally tree-like networks from stochastic block models, the communities become detectable when the $p_{\mathrm{out}}/p_{\mathrm{in}}$ ratio drops below a threshold determined by the number of clusters and the average degree \cite{Decelle:PhysRevLett}.
Correlation networks differ in two ways: they are not locally tree-like, and their link probabilities follow distributions that depend on the parameters $L$, $\rho$, $q$, and $N$ rather than taking fixed values. Despite these differences, both detectability problems share that more clusters demand a lower $p_{\mathrm{out}}/p_{\mathrm{in}}$ ratio for reliable detection.

\begin{figure*}[tb]
\centering
\includegraphics[width=\textwidth]{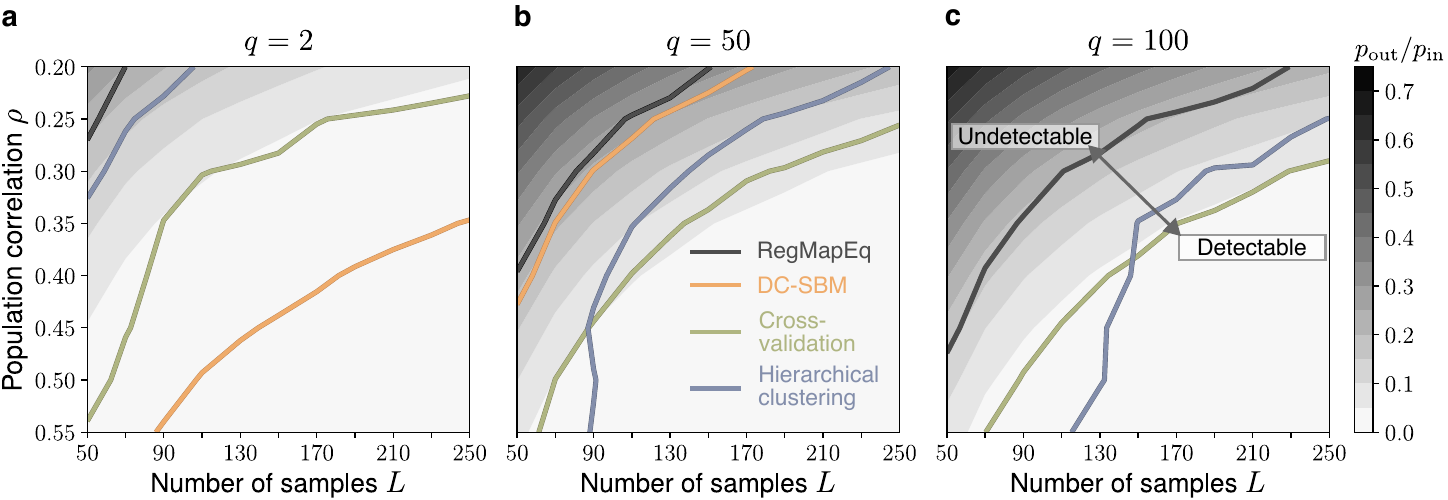}
\caption{\label{fig:AMIlim}\textbf{Detectability limits in the $L$--$\rho$ space.} Four methods compared on a network with $N=1000$ nodes and varying number of clusters $q$. Detectability defined as the line where the AMI between planted and detected partitions on average exceeds 0.95. The link ratio $p_{\mathrm{out}}/p_{\mathrm{in}}$ is shown as contour levels in the background. Cross-validation and hierarchical clustering require relatively strong signals for correct inference, and hierarchical clustering is particularly sensitive to small sample sizes. The DC-SBM faces challenges with large and dense clusters (\textbf{a}) and small clusters (\textbf{c}), where it tends to overfit and underfit respectively, but performs well in the intermediate case with $q=50$ clusters (\textbf{b}). The Regularized Map Equation (RegMapEq) pushes the detectability limit furthest into the tested high-noise regions.}
\end{figure*}

To compare methods quantitatively, we estimate a method's detectability limit as the boundary in the $L$--$\rho$ parameter space where the average AMI between the planted and inferred partitions exceeds 0.95.
Figure~\ref{fig:AMIlim} shows this limit for the evaluated methods and different numbers of clusters $q$ in a fixed-size network ($N=1000$). Hierarchical clustering struggles in noisy regimes and is particularly sensitive to small sample sizes. Its detectability limits diverge from $p_{\mathrm{out}}/p_{\mathrm{in}}$ isolines as $L$ decreases. The community detection methods show a similar tendency because they use higher thresholds to reduce false positives (Supplementary Fig.~4).
The DC-SBM underfits small clusters by merging them and overfits large and dense clusters by splitting them. The AMI never exceeds 0.95 with $q=100$ clusters (Fig.~\ref{fig:AMIlim}c), and its detectability limit for $q=2$ clusters is pushed into the low-noise region with low $p_{\mathrm{out}}/p_{\mathrm{in}}$ ratio (Fig.~\ref{fig:AMIlim}a). Only in the intermediate case with $q=50$ clusters does it outperform cross-validation and hierarchical clustering (Fig.~\ref{fig:AMIlim}b), giving it a narrow window for application.
The Regularized Map Equation's detectability limit extends into high-noise regions with relatively high $p_{\mathrm{out}}/p_{\mathrm{in}}$ ratio for all tested cluster sizes (Fig.~\ref{fig:AMIlim}), recovering the planted partition with high accuracy in the tested high-noise settings.

\subsection*{Clusters in gene co-expression data}
To assess the Regularized Map Equation's performance in practice, we analyze transcriptomic data from the Allen Human Brain Atlas, with gene expression levels measured across hundreds of brain samples (details in Methods). In this more complex and noisy setting, we compare gene cluster partitions inferred by the Regularized Map Equation and WGCNA, a widely used method in systems biology based on hierarchical clustering and soft thresholding \cite{Horvath}.

\begin{figure*}[tb]
\centering
\includegraphics[width=\textwidth]{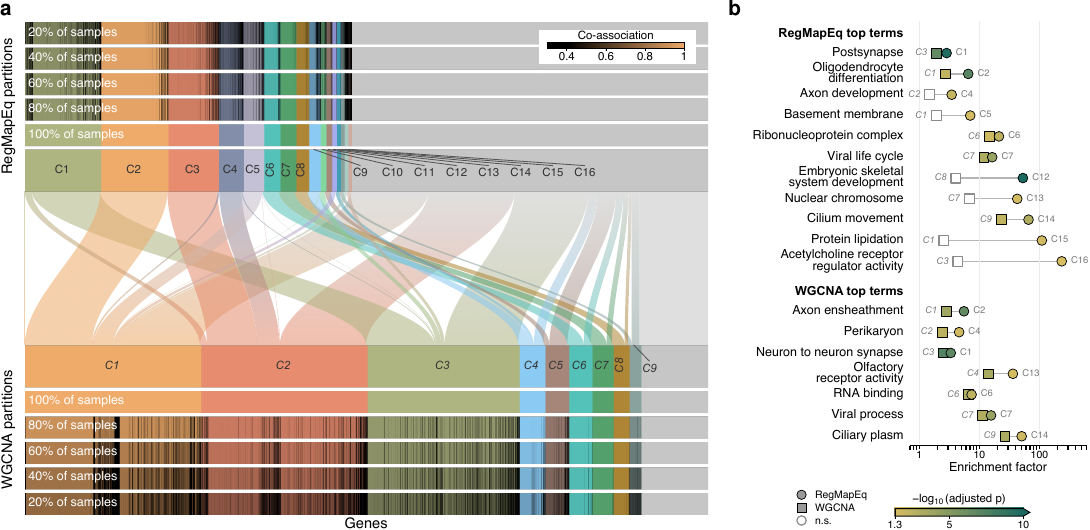} 
\caption{\label{fig:partitions}\textbf{Gene clusters from gene co-expression data.} Using data from the Allen Human Brain Atlas, we compare gene cluster partitions inferred by WGCNA and the Regularized Map Equation. (\textbf{a}) Clusters of the Regularized Map Equation (top) and WGCNA (bottom) in the full data and in subsamples with 20--80\% of the samples. Clusters are numbered by size from the left (C1, C2, \ldots), with WGCNA's clusters in italics (\textit{C1}, \textit{C2}, \ldots). 
Each WGCNA cluster takes the color of the Regularized Map Equation cluster it shares most genes with. 
Gray marks the background cluster. Darker genes have lower co-association: they stay less often with their cluster partners. Flows show the genes that two clusters share. Almost half of the genes in WGCNA's clusters (47\%) belong to the Regularized Map Equation's background cluster. With 20\% of the samples, the mean co-association is 0.77 for the Regularized Map Equation and 0.67 for WGCNA. (\textbf{b}) Top GO term of each enriched cluster, shown for that cluster and for the cluster of the other method where the term is most significant. Marker position gives the enrichment factor, marker shape the method and color the adjusted $p$-value, and open markers show terms that are not significant.
The Regularized Map Equation has the higher enrichment factor for every term, and six of its eleven top terms are not significant in WGCNA.}

\end{figure*}

WGCNA and the Regularized Map Equation produce strikingly different partitions (Fig.~\ref{fig:partitions}a). WGCNA assigns genes to few, relatively large clusters. The Regularized Map Equation identifies more and smaller clusters and places many more genes in a background cluster of weakly correlated genes, which we mark gray based on WGCNA's convention.
WGCNA instead spreads these weakly correlated genes across its other clusters, where they dilute the functional signal. Almost half (47\%) of the genes in WGCNA's clusters are weakly correlated genes that the Regularized Map Equation places in its background cluster.

To assess the biological relevance of the inferred clusters, we perform gene ontology (GO) enrichment analysis with the network genes as background (see Methods). 
With this background, a cluster is enriched only if its genes differ in function from the other genes in the network. Eleven of the Regularized Map Equation's 16 clusters and seven of WGCNA's nine clusters are enriched. 
For each enriched cluster of either method, we take its top GO term and look up the same term in all clusters of the other method. Figure~\ref{fig:partitions}b shows the result for the cluster where the term is most significant.
The Regularized Map Equation gives the higher enrichment factor for every term. All of WGCNA's top terms remain significant in the Regularized Map Equation, but six of the Regularized Map Equation's eleven top terms lose significance in WGCNA.

The oligodendrocyte cluster is a clear example. The Regularized Map Equation's second-largest cluster (C2) contains 190 genes, including the oligodendrocyte markers MBP, PLP1, MOG and MAG. WGCNA places the same genes in its largest cluster (\textit{C1}) but adds 307 other genes, 220 of them from the Regularized Map Equation's background cluster. The enrichment factor for oligodendrocyte differentiation then drops from 6.5 to 2.7, and the adjusted $p$-value rises from $7\times10^{-8}$ to 0.006. A cluster of 14 HOX genes (C12) shows the same effect more strongly. Its top term has an enrichment factor of 53 in the Regularized Map Equation (adjusted $p=7\times10^{-22}$). In WGCNA's largest cluster (\textit{C1}), which also holds these genes, the enrichment factor is 2.0 and the term is not significant (adjusted $p=0.24$). 
The lower enrichment in WGCNA is not an effect of cluster size. When we draw random sets of 190 genes from WGCNA’s largest cluster (\textit{C1}), their enrichment factor for oligodendrocyte differentiation is on average 2.7, the same as for the whole cluster, and does not reach 6.5 in any of 20,000 draws. Random sets of 14 genes, as many as in the HOX cluster, likewise never reach an enrichment factor of 53. 
The difference comes from how the methods choose cluster boundaries. By using principled model selection to determine both cluster boundaries and the number of clusters, the Regularized Map Equation keeps weakly correlated genes out of its functional clusters. 

Finally, we examine the robustness of each method by subsampling the expression data. We reduce the number of samples available for inference and measure how reproducibly each gene stays with its cluster partners across 100 subsamples with co-association analysis. A gene’s co-association is the share of subsamples in which it is in the same cluster as its cluster partners in the full data (see Methods).  
With 80\% of the samples, the two methods are about equally stable, with a mean co-association of 0.87 for the Regularized Map Equation and 0.86 for WGCNA. With fewer samples, the mean co-association drops faster for WGCNA, to 0.67 at 20\% of the samples against 0.77 for the Regularized Map Equation (Fig.~\ref{fig:partitions}a). WGCNA's large clusters loose 20--25\% of their genes at each new subsampling level. The genes that leave are mainly those that the Regularized Map Equation places in a different cluster or in its background cluster. The oligodendrocyte genes show the difference clearly. With 20\% of the samples, their mean co-association is 0.91 in the Regularized Map Equation but 0.76 in WGCNA. 
The Regularized Map Equation thus identifies functionally coherent gene clusters that stay more stable than WGCNA's clusters as data become sparse. This robustness matters when samples are limited or expensive to collect.


\section*{Discussion}
Standard approaches for clustering similarity data treat noise reduction and clustering in separate steps, targeting cluster structure only after sparsification has already constrained what clustering can find. When clusters are the objective, model selection should focus on cluster structure from the start.
We demonstrate that Bayesian community detection methods enable noise reduction and clustering in a single principled step, balancing model complexity and fit with cluster structure as the sole criterion throughout.

We evaluated two Bayesian methods grounded in the Minimum Description Length principle: the Degree-Corrected Stochastic Block Model and the Regularized Map Equation.
The DC-SBM fits a generative model to the observed network, providing a probabilistic interpretation but introducing assumptions not always valid for correlation data. In particular, it assumes conditional link independence given the block structure, an assumption that correlation networks may violate through triangle closure, potentially leading to overfitting.
The Regularized Map Equation requires no generative link model. Instead, it compresses a modular description of the network with a Bayesian prior to regularize against overfitting. In our experiments, this approach recovers planted clusters with high accuracy even under substantial noise from few samples, weak population correlation, and granular cluster structure combined.

Our application to gene co-expression data from the Allen Human Brain Atlas illustrates the practical value of this approach. The widely used WGCNA selects parameters based on scale-freeness rather than cluster structure, producing partitions with a diluted functional signal and low robustness to reduced sample sizes.
The Regularized Map Equation produces more functionally coherent clusters and maintains more stable assignments even as available samples drop substantially.

Thresholding risks discarding not only noise but also signal. We mitigate this risk by moving the sparsification decision into the clustering step itself. Clusters tolerate missing within-cluster links far better than spurious between-cluster ones. Maximizing modular compression rather than link accuracy shifts the threshold accordingly. This asymmetry points toward integrated methods that model continuous similarity data directly within a community detection framework. Such methods could push reliable cluster detection deeper into high-noise regimes.

Our results extend to more complex situations where clusters vary in size and correlation strength. The correlation distribution then becomes a superposition of contributions from the noise and from each cluster. Detectability still depends on the overlap between the noise distribution and the weakest clusters; adding clusters with stronger correlations leaves this overlap unchanged. Our detectability experiments serve as a conservative baseline: the weakest clusters determine the limit.

Our one-step approach operates directly on any similarity measure, requiring no access to the underlying sample data. The starting point can be a correlation matrix, a species similarity matrix, or a molecular distance matrix. In systems biology, neuroscience, ecology, and other fields where samples are scarce and modular organization matters, Bayesian community detection enables converting noisy similarity data into interpretable structure.

\section*{Methods}

\subsection*{Correlation distributions}
The distribution $f(r; \rho, L)$ of correlation $r$ has a known analytical expression derived by Hotelling \cite{Hotelling}:
\begin{align}
    f(r; \rho, L) =& \frac{(L-2)\Gamma(L-1)(1-\rho^2)^\frac{L-1}{2}(1-r^2)^\frac{L-4}{2}}{\sqrt{2\pi}\Gamma(L-\frac{1}{2})(1-\rho r)^\frac{2L-3}{2}}\nonumber\\ & \times {}_2F_1\left(\tfrac{1}{2}, \tfrac{1}{2}, \tfrac{2L-1}{2}, \tfrac{\rho r +1}{2}\right),
    \label{eq:dist_corr}
\end{align}
where $\Gamma(z)$ is the gamma function and ${}_2F_1(a,b,c,d)$ is Gauss' hypergeometric function. For uncorrelated data, $\rho=0$, and the distribution reduces to
\begin{align}
    f_0(r; L) =& \frac{(L-2)\Gamma(L-1)(1-r^2)^\frac{L-4}{2}}{\sqrt{2\pi}\Gamma(L-\frac{1}{2})} \nonumber\\ & \times {}_2F_1\left(\tfrac{1}{2}, \tfrac{1}{2}, \tfrac{2L-1}{2}, \tfrac{1}{2}\right).
    \label{eq:dist_zero_corr}
\end{align}

The modular structure scales the correlations into distributions $\hat{f}(r; \rho, L, q, N)$ and $\hat{f}_0(r; L, q, N)$:
\begin{equation}
\label{eq:f_hat}
    \hat{f}(r; \rho, L, q, N) = \frac{N}{N-1}\left(\frac{1}{q} - \frac{1}{N}\right)f(r; \rho, L)
\end{equation}
and
\begin{equation}
\label{eq:f0_hat}
    \hat{f}_0(r; L, q, N) = \frac{N}{N-1}\left(1-\frac{1}{q}\right)f_0(r; L),
\end{equation}
where $\frac{N}{N-1}\left(\frac{1}{q} - \frac{1}{N}\right)$ and $\frac{N}{N-1}\left(1-\frac{1}{q}\right)$ are the fractions of within- and between-cluster correlations respectively (Supplementary Note~1).

An analytical expression for the intersection $r_i$ between $\hat{f}(r; \rho, L, q, N)$ and $\hat{f}_0(r; L, q, N)$ follows from the series expansion of Gauss' hypergeometric function:
\begin{equation}
    {}_2F_1\left(\tfrac{1}{2}, \tfrac{1}{2}, \tfrac{2L-1}{2}, \tfrac{\rho r +1}{2}\right) = 1 + \mathcal{O}\left(\tfrac{1}{L}\right).
\end{equation}
This series converges fast. For large sample sizes, we can use ${}_2F_1(\frac{1}{2}, \frac{1}{2}, \frac{2L-1}{2}, \frac{\rho r +1}{2})=1$. The two distributions intersect at $r=r_i$, where $\hat{f}(r_i; \rho, L, q, N) = \hat{f}_0(r_i; L, q, N)$, equivalent to
\begin{equation}
    \left(1-\frac{1}{q}\right) = \left(\frac{1}{q} - \frac{1}{N}\right)\frac{(1-\rho^2)^\frac{L-1}{2}}{(1-\rho r_i)^\frac{2L-3}{2}}
\end{equation}
such that
\begin{equation}
    r_i = \frac{1}{\rho} - \frac{(1-q/N)^\frac{2}{2L-3}(1-\rho^2)^\frac{L-1}{2L-3}}{\rho (q-1)^\frac{2}{2L-3}},
\end{equation}
which is the threshold with the best balance between true and false positives.

\subsection*{The Regularized Map Equation}

To detect community structure in similarity data, we use the Regularized Map Equation, a Bayesian extension of the map equation framework~\cite{smiljanic1, smiljanic2}. The standard map equation identifies communities by minimizing the average per-step description length of a random walker's trajectory and works well for fully observed networks.
However, incomplete link observations can bias estimates of node visit rates and transition probabilities. Since the map equation favors partitions that shorten the description of the observed walk, the absence of links can artificially reduce the description length for highly fragmented partitions, leading to the detection of small and spurious communities~\cite{smiljanic1, smiljanic2}. Consequently, the description length estimated by the standard map equation cannot be interpreted as a reliable model selection criterion when networks are undersampled, unless one employs a cross-validation procedure~\cite{Neuman}.

The Regularized Map Equation mitigates this overfitting problem by interpreting the observed network as a finite sample from hidden transition probabilities that govern the random walk. In theory, repeated sampling under this model yields either binary links in an unweighted network or integer-valued link counts in a weighted multigraph. Rather than relying solely on these potentially sparse samples, the method uses a Bayesian rule to combine observed data with prior expectations specified by a Dirichlet distribution. This prior is uninformative and analytically tractable, reducing variance in estimated transition rates where observations are weak and producing a random walk process that is more robust to sampling noise~\cite{smiljanic1,smiljanic2}.

We construct our similarity networks from correlation values in the interval $[0,1]$. Because the weighted Regularized Map Equation is formally derived for integer-valued link counts, applying it directly to correlation weights violates the underlying multinomial sampling assumption. We therefore threshold the networks and treat them as unweighted, setting links below the threshold to zero and links above the threshold to one, and apply the unweighted Regularized Map Equation. This preserves the probabilistic interpretation of the method without introducing assumptions about fractional weights. To ensure that this modeling choice has no effect on our results, we also apply the weighted Regularized Map Equation to correlation-weighted networks and find that it produces the same detected community structure.

We use the implementation provided in Infomap, which supports regularized inference with the same computational efficiency as the standard map equation~\cite{MapEqRev}. For each evaluated threshold, we construct an unweighted and undirected network and set the prior connectivity parameter to its default value of ${\ln{N}}/{N}$~\cite{smiljanic1}.

\subsection*{Hierarchical clustering}
We perform agglomerative hierarchical clustering by using the correlation distance $1-r$ as distance between features.
We calculate the distance $d(u,v)$ between clusters $u$ and $v$ using the farthest point algorithm, such that
\begin{equation}
    d(u,v) = \max_{i,j} \mathrm{dist}(u_i, v_j)
\end{equation}
where $u_i, v_j$ are features in clusters $u, v$ respectively and $\mathrm{dist}(u_i, v_j)$ denotes the distance between features. At each level in the resulting tree-like structure we calculate the within-cluster sum of squares (WCSS), which measures the compactness of clusters, as
\begin{equation}
    \text{WCSS}(q) = \sum_{i=1}^{q} \sum_{X_j \in C_i} \| X_j - \bar{X}_i \|^2
\end{equation}
where $C_i$ denotes cluster $i$ and $\bar{X}_i$ is the centroid of cluster $i$ defined as
\begin{equation}
    \bar{X}_i = \sum_{X_j \in C_i} X_j/|C_i|.
\end{equation}
We then obtain the WCSS values as a function of the number of clusters $q$, and we cut the tree at the level where the WCSS has its maximum change in slope, as illustrated in Supplementary Fig.~1.

\subsection*{Gene co-expression data}
We use normalized microarray expression data from the Allen Human Brain Atlas \cite{Allen} (donor ID: 9861) and pre-process the data as described in ref.~\citenum{Hawrylycz}. We filter genes to retain the top 10\% most variable genes across brain samples, resulting in a set of 1934 genes and 946 samples for analysis. To simulate reduced experimental conditions, we construct gene expression matrices at four subsampling levels---80\%, 60\%, 40\% and 20\%---by randomly selecting samples without replacement. At each level, we repeat the analysis across 100 independent subsamples.

We run the Regularized Map Equation as described above and in the companion notebook \cite{RegMapEq}. For the full data and for each subsample, we scan thresholds from $\tau=0.65$ to $0.87$ in steps of 0.01 and choose the threshold $\tau^*$ with the highest compression. To find the best partition, we use 100 trials per threshold in the scan and 1000 trials for the final partition at $\tau^*$, each trial with 100 random moves. The full data gives $\tau^*=0.79$. We run WGCNA through the standard R package \cite{Langfelder} (version 1.73), following ref.~\citenum{Hawrylycz}. We build an unsigned network from Pearson correlations raised to the soft-threshold power, compute the unsigned topological overlap matrix (TOM), and cluster genes by average-linkage hierarchical clustering of 1-TOM. We define clusters with the dynamic hybrid tree cut using the defaults parameters, meaning deepSplit 1, a minimum cluster size of 20, and the PAM stage (Partitioning Around Medoids), which assigns remaining genes to the nearest cluster. We do not merge similar clusters. Without the PAM stage, 36\% of the genes in WGCNA’s clusters still belong to the Regularized Map Equation’s background cluster (47\% with it). WGCNA also chooses its soft-threshold power parameter separately for each dataset. For the full data and for each subsample, we use the package's automatic choice: the lowest power at which the scale-free fit index $R^2$ reaches 0.85. The full data gives power 8. All datasets reach this value, also with 20\% of the samples. As a check, we also run WGCNA with a fixed power of 8 in every subsample. Then 17\% of its clustered genes stay stable with 20\% of the samples, compared with 12\% when the power is chosen per subsample, so the drop in stability comes mainly from the smaller sample size.

We assess partition robustness with a co-association analysis \cite{monti2003}. For each subsampling level we cluster the 100 subsamples independently and, for every pair of genes, compute the co-association $A_{ij}$: the fraction of subsamples in which the two genes are placed in the same cluster. Genes in the background cluster are treated as unclustered. For each gene $i$ that belongs to a coherent cluster $C$ in the full-data partition, we define its stability as the mean co-association with the other members of that cluster, $a_i = \frac{1}{|C|-1}\sum_{j\in C, j\neq i} A_{ij}$. 
This measure uses pairs of genes, so it requires no matching of cluster labels across subsamples and is unaffected when clusters are relabeled or merged. In Fig. \ref{fig:partitions}a, a gene's cluster color darkens as $a_i$ decreases and is black for $a_i \leq 0.4$. To compare the methods, we report the mean co-association over all genes in clusters.

To assess biological relevance, we perform gene ontology (GO) enrichment analysis with the g:Profiler software \cite{gProfiler} through its Python package, using the genes in the network as background (1933 unique gene symbols). We correct for multiple testing with the Benjamini--Hochberg procedure and call a term enriched if its adjusted $p<0.05$. For each method and each cluster with at least 10 genes, excluding the background cluster, we compute the enrichment factor of each enriched term: the observed proportion of cluster genes in the term divided by the expected proportion. 
To compare the methods, we take the top term of each enriched cluster and look up the same term in every cluster of the other method. 
Fig.~\ref{fig:partitions}b shows the term in the cluster where it is most significant.

\section*{Data availability}
The microarray gene expression data analyzed in this study are publicly available from the Allen Human Brain Atlas at \url{https://human.brain-map.org} (donor ID: 9861). The synthetic datasets are generated as described in the Methods and can be reproduced with the code referenced below. 

\section*{Code availability}
The Regularized Map Equation is implemented in Infomap, available at \url{https://www.mapequation.org}. A companion notebook illustrating the one-step approach is available \cite{RegMapEq}. 

\bibliographystyle{naturemag}
\bibliography{refs}

\section*{Acknowledgements}
The authors thank the High Performance Computing Center North (HPC2N) at Umeå University for providing computational resources and valuable support. We also thank M.~Hawrylycz for advice on the Allen Human Brain Atlas data. M.R. was supported by the Swedish Research Council under grants 2022-06543 and 2023-03705.

\section*{Author contributions}
M.N. designed the study and performed the experiments with input from J.S. and M.R. All authors wrote the manuscript.

\section*{Competing interests}
The authors declare no competing interests.

\end{document}